\documentclass[letterpaper]{article}
\usepackage{proceed2e}
\usepackage[margin=1in]{geometry}

\usepackage{times}

\title{Structured nonlinear variable selection}

\author{
\hspace{0.5cm} Magda Gregorov\'a \hspace{2cm}
Alexandros Kalousis \hspace{2.5cm}
St\'ephane Marchand-Maillet \\
Geneva School of Business Administration, HES-SO, Switzerland
\hspace{2.2cm} University of Geneva \hspace{2cm}\\
\hspace{1cm} University of Geneva, Switzerland \hspace{5.5cm} Switzerland
}

\usepackage{mathtools}
\usepackage{amssymb}
\usepackage{amsthm}
\usepackage[round]{natbib}
\usepackage{caption}
\usepackage{paralist}

\usepackage{hyperref}
\DeclareMathOperator*{\argmin}{argmin}
\DeclareMathOperator*{\vcc}{vec}%
\usepackage{xcolor,colortbl}
\definecolor{light-gray}{gray}{0.7}
\usepackage{algorithm}
\usepackage{multirow}
\newcommand{\vc}[1]{\mathbf{#1}}
\newcommand{\mF}{\mathcal{F}}
\newcommand{\mX}{\mathcal{X}}
\newcommand{\mY}{\mathcal{Y}}
\newcommand{\mR}{\mathbb{R}}
\newcommand{\mcR}{\mathcal{R}}
\newcommand{\mN}{\mathbb{N}}
\newcommand{\bx}{\mathbf{x}}
\newcommand{\bz}{\mathbf{z}}
\newcommand{\by}{\mathbf{y}}
\newcommand{\bw}{\mathbf{w}}
\newcommand{\pa}{\pmb{\alpha}}
\newcommand{\pb}{\pmb{\beta}}
\newcommand{\po}{\pmb{\omega}}
\newcommand{\pv}{\pmb{\varphi}}
\newcommand{\pl}{\pmb{\lambda}}
\newcommand{\bg}{\mathbf{g}}
\newcommand{\bs}{\mathbf{s}}
\newcommand{\bv}{\mathbf{v}}
\newcommand{\bB}{\mathbf{B}}
\newcommand{\bK}{\mathbf{K}}
\newcommand{\bD}{\mathbf{D}}
\newcommand{\bbD}{\mathbf{\tilde{D}}}
\newcommand{\bL}{\mathbf{L}}
\newcommand{\bF}{\mathbf{F}}
\newcommand{\bQ}{\mathbf{Q}}
\newcommand{\bZ}{\mathbf{Z}}

\newcommand{\nn}{\nonumber \\}%

\newtheorem{remark}{Remark}
\newtheorem{proposition}{Proposition}

\def\tablesize{\@setsize\tablesize{8pt}\viipt\@viipt}

\begin{document}

\maketitle

\begin{abstract}
We investigate structured sparsity methods for variable selection in regression problems where the target depends nonlinearly on the inputs.
We focus on general nonlinear functions not limiting a priori the function space to additive models.
We propose two new regularizers based on partial derivatives as nonlinear equivalents of group lasso and elastic net.
We formulate the problem within the framework of learning in reproducing kernel Hilbert spaces and show how the variational problem can be reformulated into a more practical finite dimensional equivalent.
We develop a new algorithm derived from the ADMM principles that relies solely on closed forms of the proximal operators.
We explore the empirical properties of our new algorithm for Nonlinear Variable Selection based on Derivatives (NVSD) on a set of experiments and confirm favourable properties of our structured-sparsity models and the algorithm in terms of both prediction and variable selection accuracy.
\end{abstract}

\section{Introduction}
\label{sec:intro}

We are given a set of $n$ input-output pairs $\{ (\bx^i,y^i) \in (\mX \times \mY) \, : \, \mX \subseteq \mR^d, \mY \subseteq \mR, \, i \in \mN_n \}$ sampled i.i.d. according to an unknown probability measure $\rho$.
Our task is to learn a regression function $f : \mX \to \mY$ with minimal expected squared error loss $\mathcal{L}(f) = \mathbb{E} \left(y - f(\bx) \right)^2 = \int \left(y - f(\bx) \right)^2 d\rho(\bx,y)$.

We follow the standard theory of regularised learning 
where $\widehat{f}$ is learned by minimising the regularised empirical squared error loss $\mathcal{\widehat{L}}(f) = \frac{1}{n} \sum_i^n \left( y^i - f(\bx^i) \right)^2$
\begin{equation}\label{eq:RegLearning}
\widehat{f} = \argmin_{f} \, \mathcal{\widehat{L}}(f) + \tau \mcR(f) \enspace .
\end{equation}
In the above, $\mcR(f)$ is a suitable penalty typically based on some prior assumption about the function space (e.g. smoothness), and $\tau > 0$ is a suitable regularization hyper-parameter.
The principal assumption 
we consider in this paper is that the function $f$ is sparse with respect to the original input space $\mX$, that is it depends only on $l \ll d$ input variables.

Learning with variable selection is a well-established and rather well-explored problem in the case of linear models $f(\bx) = \sum_a^d x_a w_a$, e.g. \citep{Hastie2015}.
The main ideas from linear models have been successfully transferred to additive models $f(\bx) = \sum_a^d f_a(x_a)$, e.g. \citep{Ravikumar2007,Bach2009a,Koltchinskii2010,Yin2012}, or to additive models with interactions $f(\bx) = \sum_a^d f_a(x_a) + \sum_{a<b}^d f_{a,b}(x_a,x_b)$, e.g \citep{Lin2006,Tyagi2016}.

However, sparse modelling of general non-linear functions is more intricate.
A promising stream of works focuses on the use of non-linear (conditional) cross-covariance operators arising from embedding probability measures into Hilbert function spaces, e.g. \citep{Yamada2014,Chen2017}.



In this work, we follow an alternative approach proposed in \citep{Rosasco2013} based on partial derivatives and
develop new regularizers to promote structured sparsity with respect to the original input variables.
We stress that our objective here is not to learn new data representations nor learn sparse models in some latent feature space, e.g. \citep{Gurram2014}.
Nor is it to learn models sparse in the data instances (in the sense of support vectors, e.g. \citep{Chan2007}). We aim at selecting the relevant input variables, the relevant dimensions of the input vectors $\bx \in \mR^d$.

After a brief review of the regularizers used in \citep{Rosasco2013} for individual variable selection in non-linear model learning (similar in spirit to lasso \citep{Tibshirani1996}) we propose two extensions motivated by the linear structured-sparsity learning literature.
Using suitable norms of the partial derivatives we propose the non-linear versions of the group lasso \citep{Yuan2006} and the elastic net \citep{Zou2005}.

We pose our problem into the framework of learning in the reproducing kernel Hilbert space (RKHS).
We extend the representer theorem to show that the minimiser of \eqref{eq:RegLearning} with our new regularizers $\mcR(f)$ can be conveniently written as a linear combination of kernel functions and their partial derivatives evaluated over the training set.

We further propose a new reformulation of the equivalent finite dimensional learning problem, which 
allows us to develop a new algorithm (NVSD) based on the Alternating Direction Method of Multipliers (ADMM) \citep{Boyd2010}.
This is a generic algorithm that can be used (with small alterations) for all regularizers we discuss here.
At each iteration, the algorithm needs to solve a single linear problem, perform a proximal step resulting in a soft-thresholding operation, and do a simple additive update of the dual variables.
Unlike \citep{Rosasco2013}, which uses approximations of the proximal operator, our algorithm is based on proximals admitting closed forms for all the discussed regularizers, including the one suggested previously in \citep{Rosasco2013}.
Furthermore, by avoiding the approximations in the proximal step, the algorithm directly provides also the learned sparsity patterns over the training set (up to the algorithmic convergence precision).

We explore the effect of the proposed regularizers on model learning on synthetic and real-data experiments,
and confirm the superior performance of our methods in comparison to a range of baseline methods when learning structured-sparse problems.
Finally, we conclude by discussing the advantages and shortcomings of the current proposal and outline some directions for future work.


\section{Regularizers for variable selection}
\label{sec:Regularizers}

In \citep{Rosasco2013} the authors propose to use the partial derivatives of the function  with respect to the input vector dimensions $\{ \partial_a f: a \in \mN_d \}$ to construct a regularizer promoting sparsity.
The partial derivative evaluated at an input point $\partial_a f(\bx)$ is the rate of change of the function at that point with respect to $x_a$ holding the other input dimensions fixed.
Intuitively, when the function does not dependent on an input variable (input dimension $a$), its evaluations do not change with changes in the input variable: $\partial_a f(\bx) = 0$ at all points $\bx \in \mX$.
A natural measure of the size of the partial derivatives across the space $\mX$ is the $L_2$ norm
\begin{equation}\label{eq:DerivativeNorm}
|| \partial_a f ||_{L_2} = \sqrt{ \int_{\mX} | \partial_a f(\bx) |^2 \, d\rho_x(\bx)} 
\end{equation}

\begin{remark}
At this point we wish to step back and make a link to the linear models $f(\bx) = \sum_a^d \, x_a w_a$.
The partial derivatives with respect to any of the $d$ dimensions of the input vector $\bx$ are the individual elements of the $d$-dimensional parameter vector $\bw$, $\partial_a f(\bx) = w_a$, and this at every point $\bx \in \mX$.
For the linear model we thus have $|| \partial_a f ||_{L_2} = |w_a|$.
Sparsity inducing norms or constraints operating over the parameter vectors $\bw$ can therefore be seen as special cases of the same norms and constraints imposed on the partial-derivative norms \eqref{eq:DerivativeNorm}.
\end{remark}

\subsection{Sparsity inducing norms}
\label{sec:SparsityNorms}
The sparsity objective over a vector $\bv \in \mR^d$ can be cast as the minimization of the $\ell_0$ norm $|| \bv ||_0 = \# \{ a = 1,\dots,d : v_a \neq 0 \} $ which counts the number of non-zero elements of the vector.
Since it is well known from the linear sparse learning literature that finding the $\ell_0$ solutions is computationally difficult in higher dimensions (NP-hard, \citep{Weston2003}), the authors in \citep{Rosasco2013} suggest to use its tightest convex relaxation, the $\ell_1$ norm $||\bv||_1 = \sum_a^d |v_a|$.
They apply the $\ell_1$ norm over the partial-derivative norms \eqref{eq:DerivativeNorm} so that the lasso-like sparsity regularizer in \eqref{eq:RegLearning} is 
\begin{equation}\label{eq:RegL}
\mcR^L(f) = \sum_{a=1}^d || \partial_a f ||_{L_2} 
\enspace .
\end{equation}

In this paper we explore two extensions inspired by the linear sparse learning, opening the doors to many of the other sparsity and structured sparsity inducing norms that have been proposed in the abundant literature on this topic.
Namely, we focus here on the structured sparsity induced by the mixed $\ell_1/\ell_2$ norm known in the context of linear least squares as the group lasso \citep{Yuan2006}.
For a vector $\bv$ composed of $G$ groups $\bv_g$ (non-overlapping but not necessarily consecutive) with $p_g$ number of elements each, the mixed $\ell_1/\ell_2$ norm is $||\bv||_{1,2} = \sum_g^G p_g \: ||\bv_g||_2$.
The corresponding group-lasso-like regularizer to be used in \eqref{eq:RegLearning} is 
\begin{equation}\label{eq:RegGL}
\mcR^{GL}(f) = \sum_{g=1}^{G} p_g \sqrt{ \sum_{a \in g} || \partial_a f ||_{L_2}^2}
\enspace .
\end{equation}

Second, we look at the elastic net penalty proposed initially in \citep{Zou2005}.
This uses a convex combination of the $\ell_1$ and square of the $\ell_2$ norm and
has been shown to have better selection properties over the vanilla $\ell_1$ norm regularization in the presence of highly correlated features.
Unlike the $\ell_1$ penalty, the combined elastic net is also strictly convex.
The corresponding elastic-net-like regularizer to be used in \eqref{eq:RegLearning} is 
\begin{align}\label{eq:RegEN}
\mcR^{EN}(f) = & \, \mu \, \sum_{a=1}^d || \partial_a f ||_{L_2} 
+ (1-\mu) \, \sum_{a=1}^d || \partial_a f ||_{L_2}^2 , \nonumber \\
& \qquad \mu \in [0,1]
\enspace .
\end{align}

\subsection{Empirical versions of regularizers}
\label{sec:EmpiricalRegularizers}
A common problem of the regularizers introduced above is that in practice they cannot be evaluated due to the unknown probability measure $\rho_x$ on the input space $\mX$.
Therefore instead of the partial-derivative norms defined in expectation in \eqref{eq:DerivativeNorm}
\begin{equation}\label{eq:DerivativeNormL2}
|| \partial_a f ||_{L_2} = \sqrt{ \mathbb{E}  | \partial_a f(\bx) |^2 } 
\end{equation}
we use their sample estimates replacing the expectation by the training sample average
\begin{equation}\label{eq:DerivativeNorm2n}
|| \partial_a f ||_{2_n} = \sqrt{ \frac{1}{n} \sum_i^n  | \partial_a f(\bx^i) |^2 }
\enspace .
\end{equation}
This corresponds to the move from expected loss to the empirical loss introduced in section \ref{sec:intro} and is enabled by the i.i.d. sample assumptions.

In result, the regression function is learned from the empirical version of \eqref{eq:RegLearning}
\begin{equation}\label{eq:RegLearningEmpirical}
\widehat{f} = \argmin_{f \in \mF} \, \mathcal{\widehat{L}}(f) + \tau \widehat{\mcR}(f)
\enspace ,
\end{equation}
where $\widehat{\mcR}(f)$ are the empirical analogues of the regularizers \eqref{eq:RegL}, \eqref{eq:RegGL} and \eqref{eq:RegEN} replacing the population partial-derivative norms $|| \partial_a f ||_{L_2}$ by their sample estimates $|| \partial_a f ||_{2_n}$.
The function space $\mF$ is discussed next.

\section{Learning in RKHS}
\label{sec:RKHS}
In this paper, the hypothesis space $\mF$ within which we learn the function $f$ is a reproducing kernel Hilbert space (RKHS). 
We recall (e.g. \citep{Saitoh2016}) that a RKHS is a function space ${\mF}$ of real-valued functions over $\mX$ endowed with an inner product $\langle ., . \rangle_{\mF}$ and the induced norm $||.||_\mF$ that is uniquely associated with a positive semidefinite kernel $k: \mX \times \mX \to \mR$.
The kernel $k$ has the reproducing property $\langle k_{\bx}, f \rangle_{\mF} = f(\bx)$ and, in particular, $\langle k_{\bx}, k_{\bx'} \rangle_{\mF} = k(\bx,\bx')$, where $k_\bx \in \mF$ is the kernel section centred at $\bx$ such that $k_{\bx}(\bx') = k(\bx,\bx')$ for any two $\bx, \bx' \in \mX$. 
Furthermore, the space $\mF$ is the completion of the linear span of the functions $\{k_\bx : \bx \in \mX \}$.

In addition to these fairly well known properties of the RKHS and its kernel, the author in \citep{Zhou2008} has shown that if $k$ is continuous and sufficiently smooth
the kernel partial-derivative functions belong to the RKHS and have a partial-derivative reproducing property.
More specifically, we define the kernel partial-derivative function $[\partial_a k_\bx] : \mX \to \mR$ as
\begin{equation}
[\partial_a k_\bx] (\bx') = \frac{\partial}{\partial x_a} k(\bx,\bx') \quad \forall \bx,\bx' \in \mX
\enspace .
\end{equation}
The function $[\partial_a k_\bx] \in \mF$ has the reproducing property $\langle [\partial_a k_\bx] , f \rangle_{\mF} = \partial_a f (\bx)$. 
In particular $\langle [\partial_a k_\bx] , k_{\bx'} \rangle_{\mF} = \partial_a k_{\bx'} (\bx)$ and 
$\langle [\partial_a k_\bx] , [\partial_b k_{\bx'}] \rangle_{\mF} = \frac{\partial^2}{\partial x_a \partial x'_b} k (\bx, \bx')$.

\begin{remark}
Since the notation above may seem somewhat knotty at first, we invite the reader to appreciate the difference between the function $[\partial_a k_\bx]$ and the partial derivative of the kernel section with respect to the $a$th dimension $\partial_a k_{\bx}$.
Clearly, $[\partial_a k_\bx] (\bx') \neq \partial_a k_{\bx} (\bx')$ for any $\bx \neq \bx' \in \mX$.
However, due to the symmetry of the kernel we do have $[\partial_a k_{\bx}] (\bx') = \partial_a k_{\bx'} (\bx) = \frac{\partial}{\partial x_a} k(\bx,\bx')$.
\end{remark}

\subsection{Solution representation}
\label{sec:SolutionRepresentation}
The variational (infinite-dimensional) problem \eqref{eq:RegLearningEmpirical} is difficult to handle as is.
However, it has been previously shown for a multitude of RKHS learning problems that their solutions $\widehat{f}$ can be expressed as finite linear combinations of the kernel evaluations over the training data \citep{Argyriou2014a}.
This property, known as \emph{representer theorem}, renders the problems amenable to practical computations.

\begin{proposition}\label{prop:ReprTheorem}
The minimising solution $\widehat{f}$ of the variational problem 
\begin{equation}\label{eq:RegLearningEmpiricalH}
\widehat{f} = \argmin_{f \in \mF} \, \mathcal{\widehat{L}}(f) + \tau \widehat{\mcR}(f)
+ \nu ||f||^2_{\mF} 
\enspace ,
\end{equation}
where $\tau,\nu \geq 0$ and $\widehat{\mcR}(f)$ is any of the empirical versions of the three formulations \eqref{eq:RegL}, \eqref{eq:RegGL}, \eqref{eq:RegEN}
can be represented as
\begin{equation}\label{eq:Representation}
\widehat{f} = \sum_i^n \alpha_i \, k_{\bx^i} + \sum_i^n \sum_a^d \beta_{ai} \, [\partial_a k_{\bx^i}]
\enspace .
\end{equation}
\end{proposition}

The proof (available in the appendix) follows the classical approach (e.g. \citep{Scholkopf2001}) of decomposition of $\mF$ into the space spanned by the representation and its orthogonal complement.

The proposition extends the representer theorem of \citep{Rosasco2013} to the new regularizers \eqref{eq:RegGL} and \eqref{eq:RegEN}. 
Note that we included the induced Hilbert norm $||f||_\mF$ into \eqref{eq:RegLearningEmpiricalH} as a useful generalization that reduces to our original problem \eqref{eq:RegLearningEmpirical} if $\nu = 0$.
On the other hand, when $\tau = 0$ we recover a classical kernel regression problem which is known to have another simpler representation consisting just of the first term in \eqref{eq:Representation}.

\section{Algorithm}
\label{sec:Algorithm}
In this section we describe the new algorithm we developed 
to solve problem \eqref{eq:RegLearningEmpiricalH} with the three sparse regularizers introduced in section \ref{sec:Regularizers}.
The algorithm is versatile so that it requires only small alterations in specific steps to move from one regularizer to the other.
Importantly, unlike the algorithm proposed in \citep{Rosasco2013} for solving only the lasso-like problem, our algorithm does not need to rely on proximal approximations since all the proximal steps can be evaluated in closed forms.
Our algorithm also directly provides values of the partial derivatives of the learned function indicating the learned sparsity.

\subsection{Finite dimensional formulation}
\label{sec:FiniteProblem}
To be able to develop a practical algorithm we first need to reformulate the variational optimisation problem \eqref{eq:RegLearningEmpiricalH} into its finite dimensional equivalent.
For this we introduce the following objects:
the $n$-long vector $\pa = [\alpha_1, \dots, \alpha_n]^T$,
the $dn$-long vector $\pb = [\beta_{11}, \dots, \beta_{1n}, \beta_{21} \dots \beta_{dn}]^T$,
the $n \times n$ symmetric PSD kernel matrix $\bK$ such that $K_{ij} = k(\bx^i,\bx^j)$,
the $n \times n$ (non-symmetric) kernel derivative matrices $\bD^a$ and $\bbD^a, a \in \mN_d$ such that $D^a_{ij} = [\partial_a k_{\bx^i}] (\bx^j) = \partial_a k_{\bx^j} (\bx^i) = \tilde{D}^a_{ji}$,
the $n \times n$ (non-symmetric) kernel 2nd derivative matrices $\bL^{ab}, a,b \in \mN_d$ such that
$\bL^{ab}_{ij} = \frac{\partial^2}{\partial x^i_a \partial x^j_b} k(\bx^i,\bx^j) = \frac{\partial}{\partial x^j_b} [\partial_a k_{\bx^i}] (\bx^j) = \bL^{ba}_{ji}$.
Further, we need the following concatenations:
\begin{equation*}
\bD =
\begin{bmatrix}
\bD^1 \\
\dots \\
\bD^d 
\end{bmatrix}
\quad
\bL^a =
[\bL^{a1} \dots \bL^{ad}]
\quad
\bL =
\begin{bmatrix}
\bL^1 \\
\dots \\
\bL^d 
\end{bmatrix}
\end{equation*}
and specifically for the groups $g$ in $\mcR^{GL}$ the partitions
\begin{equation*}
\ddot{\bD}^g =
\begin{bmatrix}
\bD^{g_1} \\
\dots \\
\bD^{g_{p_g}} 
\end{bmatrix}
\quad
\ddot{\bL}^g =
\begin{bmatrix}
\bL^{g_1} \\
\dots \\
\bL^{g_{p_g}} 
\end{bmatrix}
\enspace ,
\end{equation*}
where the subscripts $g_i$ are the corresponding indexes of the input dimensions.

\begin{proposition}\label{prop:Fin1}
The variational problem \eqref{eq:RegLearningEmpiricalH} is equivalent to the finite dimensional problem
\begin{align}\label{eq:FiniteProblem}
\argmin_{\pa,\pb} \, \mathcal{J}1 (\pa,\pb)
+ \tau \mathcal{J}2 (\pa,\pb)
+ \nu \mathcal{J}3 (\pa,\pb), 
\end{align}
where
\begin{align*}
 & \mathcal{J}1 (\pa,\pb) = \frac{1}{n} || \by - \bK \pa - \bD^T \pb ||_2^2 \\
 \mcR^L: \ & \mathcal{J}2 (\pa,\pb) = \frac{1}{\sqrt{n}} \sum_a^d ||\bD^a \pa + \bL^a \pb ||_2 \\
 \mcR^{GL}:  \ & \mathcal{J}2 (\pa,\pb) = \frac{1}{\sqrt{n}} \sum_g^G p_g \, ||\ddot{\bD}^g \, \pa + \ddot{\bL}^g \, \pb ||_2 \nonumber \\
 \mcR^{EN}: \ & \mathcal{J}2 (\pa,\pb) = \frac{\mu}{\sqrt{n}} \sum_a^d ||\bD^a \pa + \bL^a \pb ||_2 \\
 & \qquad \qquad {} + \frac{1-\mu}{n} \sum_a^d ||\bD^a \pa + \bL^a \pb ||^2_2 \\
 & \mathcal{J}3 (\pa,\pb) = \pa^T \bK \pa + 2 \pa^T \bD^T \pb + \pb^T \bL \pb
\end{align*}
\end{proposition}

The proof (available in the appendix) is based on the finite dimensional representation \eqref{eq:Representation} of the minimising function, and the kernel and derivative reproducing properties stated in section \ref{sec:RKHS}.

The problem reformulation \eqref{eq:FiniteProblem} is instructive in terms of observing the roles of the kernel and the derivative matrices and is reminiscent of the classical finite dimensional reformulation of Hilbert-norm regularised least squares.
However, for the development of our algorithm we derive a more convenient equivalent form. 
\begin{proposition}\label{prop:Fin2}
The variational problem \eqref{eq:RegLearningEmpiricalH} is equivalent to the finite dimensional problem
\begin{align}\label{eq:FiniteProblemOmega}
\argmin_{\po} 
\frac{1}{n} || \by - \bF \po ||_2^2
+ \tau \mathcal{J} (\po)
+ \nu \, \po^T \bQ \, \po, 
\end{align}
where
\begin{align*}
\mcR^L: \ & \mathcal{J} (\po) = \frac{1}{\sqrt{n}} \sum_a^d ||\bZ^a \po ||_2\ \\
\mcR^{GL}: \ & \mathcal{J} (\po) = \frac{1}{\sqrt{n}} \sum_g^G p_g \,  ||\ddot{\bZ}^g \, \po||_2 \\
\mcR^{EN}: \ & \mathcal{J} (\po) = \frac{\mu}{\sqrt{n}} \sum_a^d ||\bZ^a \po||_2 
+ \frac{1-\mu}{n} \sum_a^d ||\bZ^a \po||^2_2  
\enspace ,
\end{align*}
with
\begin{equation*}
\po = 
\begin{bmatrix}
\pa \\
\pb
\end{bmatrix}
\quad
\begin{matrix}
\bF = [\bK \bD^T] \\
\bZ^a = [\bD^a \bL^a] \\
\ddot{\bZ}^g = [\ddot{\bD}^g \ddot{\bL}^g] 
\end{matrix}
\quad
\bQ = 
\begin{bmatrix}
\bK & \vc{0} \\
2 \bD & \bL
\end{bmatrix}
\end{equation*}
\end{proposition}

The proof is trivial using \eqref{eq:FiniteProblem} as an intermediate step.

\subsection{Development of generic algorithm}
\label{sec:DevelopmentOfAlgorithm}
Problem \eqref{eq:FiniteProblemOmega} is convex though its middle part $\mathcal{J} (\po)$ is non-differentiable for all three discussed regularizers. 
Indeed, it is the singularities of the norms at zero points that yield the sparse solutions.
A popular approach for solving convex non-differentiable problems is the proximal gradient descent \citep{Parikh2013}.
At every step it requires evaluating the proximal operator defined for any function $f: \mR^m \to \mR^m$ and any vector $\bv \in \mR^m$ as 
\begin{equation}\label{eq:ProximalDef}
prox_{f} ( \bv ) = \argmin_{\bx} f(\bx) + \frac{1}{2} || \bx - \bv ||_2^2
\enspace .
\end{equation}
However, proximal operators for the functions $ \mathcal{J} $ in \eqref{eq:FiniteProblemOmega} do not have closed forms or fast methods for solving which makes the proximal gradient descent algorithm difficult to use.

We therefore propose to introduce a linearizing change of variables $\bZ^a \po = \pv_a$  and cast the problem in a form amenable for the ADMM method \citep{Boyd2010}
\begin{gather}
\min \ \mathcal{E} (\po)  
+ \tau \, \mathcal{I} (\pv), 
\quad \text{s.t. } \, \bZ \, \po - \pv = 0 
\enspace .
\end{gather}
In the above 
\begin{equation*}
\pv = 
\begin{bmatrix}
\pv_1 \\
\dots \\
\pv_d
\end{bmatrix}
\qquad
\bZ = 
\begin{bmatrix}
\bZ^1 \\
\dots \\
\bZ^d
\end{bmatrix}
\enspace ,
\end{equation*}
(or concatenation of the double-dot version for the group structure), 
$\mathcal{E} : \mR^{n+nd} \to \mR$ is the convex differentiable function
\begin{equation*}
\mathcal{E} (\po) = \frac{1}{n} || \by - \bF \po ||_2^2
+ \nu \, \po^T \bQ \, \po
\enspace ,
\end{equation*}
and 
$\mathcal{I} : \mR^{nd} \to \mR$ is the convex non-differentiable function corresponding to each regularizer such that $\mathcal{I}(\pv) = \mathcal{J}(\po)$ for every $\bZ \, \po = \pv$.

At each iteration the ADMM algorithm consists of the following three update steps (the standard approach of augmented Lagrangian with $\pl$ as the scaled dual variable
and $\kappa$ as the step size):
\begin{align*}
S1: \ \po^{(k+1)} & = \argmin_{\po} \ \mathcal{E} (\po)
+ \frac{\kappa}{2} \, ||\bZ \, \po - \pv^{(k)} + \pl^{(k)} ||_2^2 \\
S2: \ \pv^{(k+1)} & = \argmin_{\pv} \tau \mathcal{I}(\pv) + \frac{\kappa}{2} ||\bZ \, \po^{(k+1)} - \pv + \pl^{(k)} ||_2^2 \\
S3: \ \pl^{(k+1)} & = \pl^{(k)} + \bZ \, \po^{(k+1)} - \pv^{(k+1)}
\end{align*}

The first step $S1$ is a convex quadratic problem with a closed form solution
\begin{multline*}
S1: \ (\nu \bQ + \nu \bQ^T + 2 n^{-1} \bF^T \bF + \kappa \bZ^T \bZ) \, \po^{(k+1)} = \\
2 n^{-1} \bF^T \by + \kappa \bZ^T (\pv^{(k)} - \pl^{(k)})
\end{multline*}

By comparing with \eqref{eq:ProximalDef} we observe that the second step $S2$ is a proximal update.
The advantage of our problem reformulation and our algorithm is that this has a closed form for all the three discussed regularizers.
\begin{proposition}
The proximal problem in step $S2$ is decomposable by the $d$ partitions of vector $\pv$ (or $G$ partition in case of the group structure) and the minimising solution is
\begin{multline*}
\mcR^L: \ \pv_a^{(k+1)} =  \\
(\bZ^a \, \po^{(k+1)} + \pl_a^{(k)}) 
\left(1 - \frac{\tau}{\kappa \sqrt{n} ||\bZ^a \, \po^{(k+1)} + \pl_a^{(k)}||_2} \right)_+ 
\end{multline*}
\begin{multline*}
\mcR^{GL}: \ \pv_g^{(k+1)} =  \\
(\ddot{\bZ}^g \, \po^{(k+1)} + \ddot{\pl}_g^{(k)}) 
\left(1 - \frac{\tau \, p_g}{\kappa \sqrt{n} ||\bZ^g \, \po^{(k+1)} + \pl_g^{(k)}||_2} \right)_+ 
\end{multline*}
\begin{multline*}
\mcR^{EN}: \ \pv_a^{(k+1)} =  \\
\frac{\bZ^a \, \po^{(k+1)} + \pl_a^{(k)}}{2 \tau (1-\mu)/ (\kappa n) + 1}
\left(1 - \frac{\tau \mu}{\kappa \sqrt{n} ||\bZ^a \, \po^{(k+1)} + \pl_a^{(k)}||_2} \right)_+ 
\end{multline*}
Here $(v)_+ = \min(0,v)$ is the thresholding operator.
\end{proposition}

The decomposability comes from the additive structure of $\mathcal{I}$.
The derivation follows similar techniques as used for classical $\ell_1$ and $\ell_2$ proximals.%
\footnote{For $\mcR^{EN}$ it is more practical to add the quadratic term into $\mathcal{E} (\po)$ in $S1$ and use the corresponding scaled version of the $\mcR^{L}$ proximal in $S2$.}

\subsection{Practical implementation}
\label{sec:Practical implementation}
In practice, the $\bQ, \bF$ and $\bZ$ matrices are precomputed in a preprocessing step and passed onto the algorithm as inputs.
The matrices are directly computable using the kernel function $k$ and its first and second order derivatives evaluated at the training points (following the matrix definitions introduced in section \ref{sec:FiniteProblem}).

The algorithm converges to a global minimum %
by the standard properties of ADMM.
In our implementation (available at \url{https://bitbucket.org/dmmlgeneva/nvsd_uai2018/})
we follow a simple updating rule \cite[sec. 3.4.1]{Boyd2010} for the step size $\kappa$.
We use inexact minimization for the most expensive step $S1$, gradually increasing the number of steepest descent steps, each with complexity $\mathcal{O} \left( (nd)^2 \right)$.

Furthermore, we use $S2$ to get the values of the training sample partial-derivative norms defined in equation \eqref{eq:DerivativeNorm2n} as $||\partial_a f^{(k)} ||_{2_n} = ||\pv_a^{(k)}||_2/\sqrt{n}$. 
The sparsity pattern is obtained by examining for which of the dimensions $a \in \mN_d$ the norm is zero $||\partial_a f^{(k)} ||_{2_n} = 0$.

\section{Empirical evaluation}
\label{sec:EmpiricalEvaluation}

We conducted a set of synthetic and real-data experiments to document the efficacy of our structured-sparsity methods and the new algorithm under controlled and more realistic conditions.
We compare our methods NVSD(L), NVSD(GL) and NVSD(EN) in terms of their predictive accuracy and their selection ability to the simple (non-sparse) kernel regularised least squares (Krls), to the sparse additive model (SpAM) of \citep{Ravikumar2007}, to the non-linear cross-covariance-based method using the Hilbert Schmidt Independence Criterion in a lasso-like manner (HSIC) of \citep{Yamada2014}, and to the derivative-based lasso-like method (Denovas) of \citep{Rosasco2013}.\footnote{For HISC and Denovas we used the author's code, for SpAM the R implementation of \cite{Zhao2014}. For all algorithms we kept the default settings.}
We compared also to simple mean and linear sparse and non-sparse models. All of these performed considerably worse than the non-linear models and therefore are not listed in the summary results.
For all the sparse kernel methods we consider a two-step debiasing procedure based on variable selection via the base algorithm followed by a simple kernel regularised least squares on the selected variables.\footnote{This is native to Denovas and necessary for HSIC which otherwise does not produce a predictive model.}

\subsection{Synthetic experiments}
\label{sec:SyntheticExperiments}

We motivate each synthetic experiment by a realistic story-line and explain the data generating process here below.
In all the synthetic experiments we fix the input dimension to $d=18$ with only 6 input variables $\{1,2,3,7,8,9\}$ relevant for the model and the other 12 irrelevant.

\begin{description}
\item[\textbf{E1}] In the first experiment we focus on the NVSD(GL) which assumes the input variables can be grouped a priori by some domain knowledge (e.g. each group describes a \emph{type} of input data such as a different biological process) and the groups are expected to be completely in or out of the model.
The input variables are generated independently from a standard normal distribution and they are grouped by three into 6 groups. The output is generated from the 1st and the 3rd group as
\begin{equation*}
y = \sum_{i=1}^3 \sum_{j=i}^3 \sum_{k=j}^3 x_i x_j x_k + \sum_{q=7}^9 \sum_{r=q}^9 \sum_{s=r}^9 x_q x_r x_s + \epsilon \enspace ,
\end{equation*}
with $\epsilon \sim N(0,0.01)$.
For learning we fix the kernel to 3rd order polynomial.
\item[\textbf{E2}] In the second experiment we do not assume any a priori grouping of the variables. Instead some of the variables are strongly correlated (perhaps relating to a single phenomenon), a case for NVSD(EN).
The input variables are generated similarly as in E1 but with the pairs $\{1,7\}, \{2,8\}$ and $\{3,9\}$ strongly correlated (Pearson's population correlation coefficient $0.95$). The remaining (irrelevant) input variables are also pair-wise correlated and the output is generated as
\begin{equation*}
y = \sum_{i,j,k=1}^3 x_i x_j x_k + \sum_{q,r,s=7}^9 x_q x_r x_s + \epsilon \enspace ,
\end{equation*}
with $\epsilon \sim N(0,0.01)$.
For learning we fix the kernel to 3rd order polynomial.
\item[\textbf{E3}] In the third experiment we assume the inputs are noisy measurements of some true phenomenon (e.g. repeated measurements, measurements from multiple laboratories) for which there is no reason to prefer one over the other in the model.
We first generate the true data $z_i \sim N(0,1), i=1,\ldots,6$ and use these to generate the outputs as 
\begin{equation*}
y = 10(z_1^2 + z_3^2) e^{-2(z_1^2 + z_3^2)} + \epsilon \enspace ,
\end{equation*}
with $\epsilon \sim N(0,0.01)$.
We then generate the noisy measurements that will be used as inputs for the learning: for each $z_i$ we create three noisy measurements $x_{ij} = z_i + N(0,0.1), j=1,2,3$ (a group for the NVSD(GL) method); the input vector is the concatenation of all $x_{ij}$ so that from the 18 long concatenated input vector $\bx$ again only the set $\{1,2,3,7,8,9\}$ of the dimensions is relevant for predicting the output $y$.
For learning we fix the kernel to Gaussian with width $\sigma=4$.
\end{description}

\begin{remark}
In all the synthetic experiments we use the same experimental protocol. 
We split the data into train sets varying the size in $n = \{30, 50, 70, 90, 110\}$, a validation set of length 1000, and a test set of length 1000.
We train the models over the train sets and use the validation set to select the regularization hyper-parameters (and therefore the models) based on the minimal validation MSE.
We use dense grids of 50 points for the $\tau$ search (automatically established by the algorithm) and 5 points grid for $\mu \in \{0.1,\dots,0.9\}$.
Complete settings (also for the baseline methods) are detailed in the replication files publicly available at \url{https://bitbucket.org/dmmlgeneva/nvsd_uai2018/}.
\end{remark}

\renewcommand{\tabcolsep}{4pt}
\begin{table}[h!]
\begin{small}
\begin{center}
\caption{Results of synthetic experiments}\label{tab:SynRes}
\begin{tabular}{ c c l | c |  c |  c |  c | c}
\hline & & & & & & \\ [-1.5ex] 
\multicolumn{3}{c|}{Train size} & 30 & 50 & 70 & 90 & 110 \\
\hline & & & & & & \\ [-1.5ex] 
E1 &
\parbox[t]{1mm}{\multirow{6}{*}{\rotatebox{90}{RMSE}}} &
Krls & 12.79 & 11.66 & 10.99 & 10.43 & 9.80 \\ 
& & SpAM & 11.41 & 9.47 & 8.66 & 8.22 & 7.75 \\ 
& & HSIC & 11.37 & 10.00 & 8.58 & 7.28 & 5.68 \\ 
& & Denovas  & 11.66 & 10.87 & 12.37 & 13.28 & 11.78 \\ 
& & NVSD(L) & 11.55 & 10.22 & 9.36 & 7.90 & 7.13 \\ 
& & NVSD(GL) & \textbf{\underline{9.92}} & \textbf{\underline{7.89}} & \textbf{\underline{6.34}} & \textbf{\underline{1.94}} & \textbf{\underline{2.41}} \\ 

\hline & & & & & & \\ [-1.5ex] 
& \parbox[t]{1mm}{\multirow{6}{*}{\rotatebox{90}{Selection error}}} &
Krls & 0.67 & 0.67 & 0.67 & 0.67 & 0.67 \\ 
& & SpAM & 0.54 & 0.56 & 0.59 & 0.57 & 0.58 \\ 
& & HSIC & 0.50 & 0.48 & 0.42 & 0.35 & 0.32 \\ 
& & Denovas  & 0.49 & 0.50 & 0.53 & 0.67 & 0.73 \\ 
& & NVSD(L) & 0.49 & 0.47 & 0.48 & 0.39 & 0.32 \\ 
& & NVSD(GL) & \textbf{\underline{0.28}} & \textbf{\underline{0.24}} & \textbf{\underline{0.22}} & \textbf{\underline{0.05}} & \textbf{\underline{0.11}} \\ 

\hline
\hline & & & & & & \\ [-1.5ex] 
E2 &
\parbox[t]{1mm}{\multirow{6}{*}{\rotatebox{90}{RMSE}}} &
Krls & 27.69 & 24.83 & 22.53 & 19.14 & 18.04 \\ 
& & SpAM & 31.24 & 29.21 & 29.25 & 27.11 & 26.03 \\ 
& & HSIC & 21.74 & 15.50 & 12.02 & 9.42 & 7.67 \\ 
& & Denovas  & 24.23 & 34.33 & 17.51 & 8.89 & 11.20 \\ 
& & NVSD(L) & 21.24 & 16.59 & 11.79 & 8.61 & 7.35 \\ 
& & NVSD(EN) & \textbf{\underline{17.53}} & \textbf{\underline{10.05}} & \textbf{\underline{5.67}} & \textbf{\underline{4.29}} & \textbf{\underline{3.29}} \\ 

\hline & & & & & &\\ [-1.5ex] 
& \parbox[t]{1mm}{\multirow{6}{*}{\rotatebox{90}{Selection error}}} &
Krls & 0.67 & 0.67 & 0.67 & 0.67 & 0.67 \\ 
& & SpAM & 0.57 & 0.55 & 0.49 & 0.52 & 0.46 \\ 
& & HSIC & 0.52 & 0.42 & 0.42 & 0.35 & 0.32 \\ 
& & Denovas  & 0.46 & 0.54 & 0.40 & 0.30 & 0.26 \\ 
& & NVSD(L) & 0.46 & 0.43 & 0.36 & 0.31 & 0.29 \\ 
& & NVSD(EN) & \textbf{\underline{0.35}} & \textbf{\underline{0.20}} & \textbf{\underline{0.14}} & \textbf{\underline{0.09}} & \textbf{\underline{0.08}} \\

\hline
\hline & & & & & & \\ [-1.5ex] 
E3 &
\parbox[t]{1mm}{\multirow{7}{*}{\rotatebox{90}{RMSE}}} &
Krls & 0.65 & 0.55 & 0.54 & 0.53 & 0.50 \\ 
& & SpAM & 0.51 & 0.49 & 0.47 & 0.47 & 0.46 \\ 
& & HSIC & 0.52 & 0.47 & 0.45 & 0.44 & 0.43 \\ 
& & Denovas  & 0.55 & 0.51 & 0.50 & 0.51 & 0.50 \\ 
& & NVSD(L) & 0.51 & 0.44 & 0.44 & 0.41 & 0.34 \\ 
& & NVSD(GL) & 0.51 & \textbf{\underline{0.41}} & \textbf{\underline{0.39}} & \textbf{\underline{0.33}} & 0.31 \\ 
& & NVSD(EN) & \textbf{0.50} & 0.43 & 0.42 & \underline{0.36} & \textbf{\underline{0.30}} \\ 

\hline & & & & & &\\ [-1.5ex] 
& \parbox[t]{1mm}{\multirow{7}{*}{\rotatebox{90}{Selection error}}} &
Krls & 0.67 & 0.67 & 0.67 & 0.67 & 0.67 \\ 
& & SpAM & 0.65 & 0.61 & 0.60 & 0.58 & 0.59 \\ 
& & HSIC & 0.59 & 0.51 & 0.53 & 0.47 & 0.44 \\ 
& & Denovas  & 0.49 & 0.45 & 0.47 & 0.45 & 0.41 \\ 
& & NVSD(L) & 0.33 & 0.30 & 0.40 & 0.34 & 0.23 \\ 
& & NVSD(GL) & \textbf{\underline{0.26}} & \textbf{\underline{0.20}} & \textbf{\underline{0.24}} & \textbf{\underline{0.15}} & \textbf{\underline{0.14}} \\ 
& & NVSD(EN) & 0.30 & 0.33 & 0.35 & \underline{0.25} & \underline{0.16} \\ 

\hline
\end{tabular}
\end{center}
Best results in bold; underlined when structured-sparsity methods significantly better than all other methods using Wilcoxon signed-rank test at 5\% significance level.
\end{small}
\end{table}

We report the average results across 50 independent replications of the experiments in table \ref{tab:SynRes}.
We measure the prediction accuracy by the root mean squared error (RMSE) over the test sets and the selection accuracy by the Tanimoto distance between the true sparsity and the learned sparsity patterns (section \ref{sec:Practical implementation}).

Our structured-sparsity methods clearly outperform all the non-structured sparse learning  methods achieving better prediction accuracy based on more precise variable selection, typically with statistically significant differences.
Also, the prediction and selection accuracy generally increases (errors decrease) for larger training sample sizes suggesting our methods are well-behaved in terms of the standard statistical learning paradigms.
In the E3 experiment, NVSD(GL) performs the best having the benefit of the prior knowledge of the variable groupings. 
Remarkably, NVSD(EN) follows closely after even without such prior information, learning about the groups of correlated variables from the data when building the model.

\begin{figure}[h!]
\centering
\includegraphics[width=0.9\columnwidth]{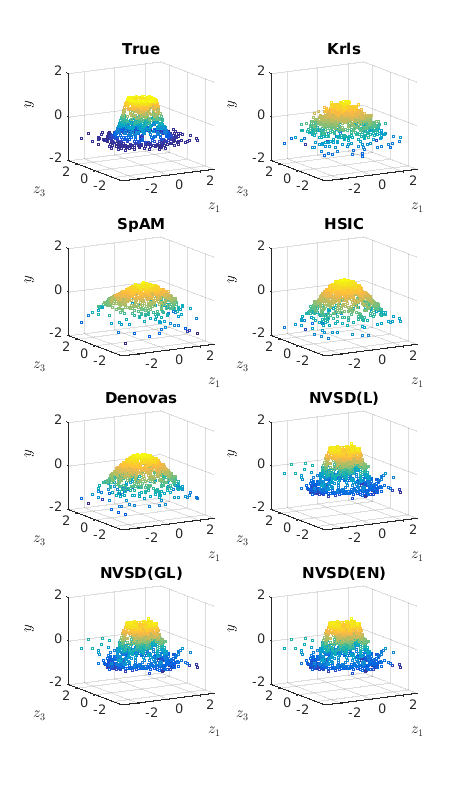}
\vspace{-2em}
\caption{Predictions for the E3 experiment over the test data. We picked an example for the model trained with 110 instances (the 17th replication) which illustrates well the advantage our NVSD methods have over the baselines in capturing the True complex non-linear structure.}\label{fig:cmRLvsDenovas}
\end{figure}

Krls can only learn full models and therefore performs rather poorly on these by-construction sparse problems.
From the other three baselines, HSIC typically achieves the second best results (after our NVSD methods).
SpAM is not particularly suitable for the non-additive structures of our experiments.
Finally, in all the experiments our NVSD(L) outperforms Denovas though they share the same lasso-like problem formulation.
We attribute this to our new algorithm developed in section \ref{sec:Algorithm}
which, unlike Denovas, does not rely on approximations of the proximal operators.

\subsection{Real-data experiments}
\label{sec:RealExperiments}

For the real-data experiments we used a collection of regression datasets from UCI \citep{Lichman2013} and LIACC\footnote{http://www.dcc.fc.up.pt/$\sim$ltorgo/Regression/DataSets.html} repositories listed in table \ref{tab:RealDes}.

\begin{table}[h!]
\begin{small}
\begin{center}
\caption{Real datasets desription}\label{tab:RealDes}
\begin{tabular}{c | c |  c |  c |  c }
\hline & & & & \\ [-1.5ex] 
Code & Name & Inputs & Test Size & Source \\
\hline & & & & \\ [-1.5ex] 
AI & Airfoil Self Noise & 5 & 700 & UCI \\ 
BH & Boston Housing & 10 & 200 & UCI \\
CC & Concrete Compressive & 8 & 450 & UCI \\
EN & Energy Efficiency & 8 & 300 & UCI \\
CP & Computer Activity & 21 & 1000 & LIACC \\
EL & F16 Elevators & 17 & 1000 & LIACC \\
KN & Kynematics & 8 & 1000 & LIACC \\
\hline
\end{tabular}
\end{center}
\end{small}
\end{table}

We report the average results across 50 replications of the experiments in tables \ref{tab:RealResE} and \ref{tab:RealResG}.
We use RMSE over the test data for measuring the prediction accuracy.
For the real datasets we do not know the ground-truth sparsity patterns.
Instead of measuring the selection error we therefore count the number of input variables selected by each method.
Krls has no selection ability, its support size is hence equal to the total number of input variables in each problem.

\begin{remark} We followed similar experimental protocol as for the synthetic experiments.
We fixed the training sample size for all experiments to 100 instances and used 200-1000 instances for the validation and test sets (depending on the total number of available observations).
We pre-processed the data by normalizing the inputs and centering the outputs.
For all the experiments we used a Gaussian kernel with the width set to the median distance calculated over the nearest 20 neighbours, and the 3rd order polynomial kernel.
With the exception of the EN dataset, the Gaussian kernel yielded better results and was therefore kept for the final evaluation.
Full details of the settings can be found in the replication files publicly available at \url{https://bitbucket.org/dmmlgeneva/nvsd_uai2018/}.
\end{remark}

Results in table \ref{tab:RealResE} are for the original data for which we have no prior knowledge about possible variable groupings.
Therefore we only use the non-structured methods and our NVSD(EN) that do not rely on any such prior information.

\begin{table}[h!]
\begin{small}
\begin{center}
\caption{Results of real-data experiments}\label{tab:RealResE}
\begin{tabular}{c l | c |  c |  c |  c | c}
\hline & & & & & \\ [-1.5ex] 
\multicolumn{2}{c|}{Experiment} & BH & CP & CC & EN & EL \\
\hline & & & & & \\ [-1.5ex] 
\parbox[t]{1mm}{\multirow{6}{*}{\rotatebox{90}{RMSE}}} &
Krls & 4.00 & 12.27 &	8.70 & 1.83 & 5.10\\
& SpAM & 4.33 & $\sim$ & 12.70 & $\sim$ & $\sim$ \\
& HSIC & 4.02 & 9.39 & 8.73 & \textbf{1.19} & 9.07 \\
& Denovas & 4.02 & 9.21 & 12.07 & 3.02 & 6.01 \\
& NVSD(L) & 3.96 & 8.43 & 	\textbf{8.67} & 1.50 & \underline{4.81} \\
& NVSD(EN) & \textbf{\underline{3.93}} & \textbf{7.88} & 8.70 & 1.20 & \textbf{\underline{4.67}} \\
\hline & & & & & \\ [-1.5ex] 
\parbox[t]{1mm}{\multirow{6}{*}{\rotatebox{90}{Support size}}} &
Krls & 10.00 & 21.00 & 8.00 & 8.00 & 17.00 \\
& SpAM & 9.00 & $\sim$ & 2.82 & $\sim$ & $\sim$ \\
& HSIC & 6.12 & 8.26 & 5.88 & 5.08 & 0.00 \\
& Denovas & 8.80 & 4.76 & 4.38 & 4.96 & 10.52 \\
& NVSD(L) & 8.20 & 3.78 & 7.36 & 7.26 & 14.06 \\
& NVSD(EN) & 8.06 & 4.58 & 7.98 & 6.66 & 13.00 \\
\hline
\end{tabular}
\end{center}
Best results in bold; underlined when NVSD methods significantly better than all the baselines using Wilcoxon signed-rank test at 5\% significance level.
For several experiments SpAM finished with errors.
\end{small}
\end{table}

Our NVSD methods learned sparse non-linear models achieving better or comparable results than the baselines in 4 out of the 5 experiments (BH, CP, EN, EL).
For CC reducing the number of input dimensions does not seem to bring any advantages and the methods tend to learn full models.
For several experiments SpAM finished with errors and therefore the results in the table are missing.

To explore the performance and benefits of NVSD(GL) method we had to construct variable groups that could potentially help the model learning. 
We adopted two strategies:
\begin{compactenum}
\item For CP and EL datasets we constructed the groups based on the NVSD(EN) results.
For CP we grouped together the 5 most often selected variables across the 50 replications of the experiment and created 3 other groups from the remaining variables. For EL we created five groups by 3-4 elements putting together variables with similar frequencies of occurrence in the support of the learned NVSD(EN) models over the 50 replications.
\item For AI, CC, and KN datasets we doubled the original input data dimensions by complementing the input data by a copy of each input variable with permuted instance order.
We then constructed two groups, the first over the original data, the second over the permuted copy.
\end{compactenum}

\begin{table}[h!]
\begin{small}
\begin{center}
\caption{Results of real-data experiments with groups}\label{tab:RealResG}
\begin{tabular}{c l | c |  c |  c |  c | c}
\hline & & & & & \\ [-1.5ex] 
\multicolumn{2}{c|}{Experiment} & AI & CP & CC & KN & EL \\
\hline & & & & & \\ [-1.5ex] 
\parbox[t]{1mm}{\multirow{6}{*}{\rotatebox{90}{RMSE}}} &
Krls & 5.08 & 12.27 &	10.34 & 2.07 & 5.10\\
& SpAM & $\sim$ & $\sim$ & 13.31 & 2.20 & $\sim$ \\
& HSIC & 4.64 & 9.39 & 9.29 & 2.05 & 9.07 \\
& Denovas & 5.12 & 9.21 & 11.49 & 2.10 & 6.01\\
& NVSD(L) & \underline{4.45} & 8.43 & 9.58 & 2.03 & \underline{4.81} \\
& NVSD(GL) & \textbf{\underline{4.16}} & \textbf{\underline{7.43}} & \textbf{\underline{8.79}} & \textbf{\underline{1.96}} & \textbf{\underline{4.76}} \\
\hline & & & & & \\ [-1.5ex] 
\parbox[t]{1mm}{\multirow{6}{*}{\rotatebox{90}{Support size}}} &
Krls & 10.00 & 21.00 & 16.00 & 16.00 & 17.00 \\
& SpAM & $\sim$ & $\sim$ & 2.60 & 11.32 & $\sim$ \\
& HSIC & 5.08 & 8.26 & 6.16 & 11.82 & 0.00 \\
& Denovas & 5.94 & 4.76 & 6.96 & 9.72 & 10.52 \\
& NVSD(L) & 4.76 & 3.78 & 8.16 & 13.58 & 14.06 \\
& NVSD(GL) & 5.00 & 5.84 & 8.00 & 11.84 & 13.82 \\
\hline
\end{tabular}
\end{center}
Best results in bold; underlined when NVSD methods significantly better than all the baselines using Wilcoxon signed-rank test at 5\% significance level.
For several experiments SpAM finished with errors.
\end{small}
\end{table}

Table \ref{tab:RealResG} confirms that our NVSD(GL) is able to use the grouping information based on prior knowledge to select better, more relevant subset of variables than the non-structured baselines.
Thanks to this it achieves significantly better prediction accuracy in all the experiments.


\section{Conclusions and future work}
\label{sec:Conclusions}

In this work we addressed the problem of variable selection in non-linear regression problems.
We followed up from the work of \cite{Rosasco2013} arguing for the use of partial derivatives as an indication of the pertinence of an input variable for the model.
Extending the existing work, we proposed two new derivative-based regularizers for learning with structured sparsity in non-linear regression similar in spirit to the linear elastic net and group lasso.

After posing the problems into the framework of RKHS learning, we designed a new NVSD algorithm for solving these. 
Unlike the previously proposed Denovas our new algorithm does not rely on proximal approximations. 
This is most likely the main reason why our NVSD(L) method achieved systematically better predictive performance than Denovas on a broad set of experiments.
We also empirically demonstrated the advantages our structured sparsity methods NVSD(GL) and NVSD(EN) bring for learning tasks with a priori known group structures or correlation in the inputs.


These promising results point to questions requiring further attention:

Our NVSD algorithm achieves better results  in terms of prediction accuracy than Denovas, however, at the cost of longer training times.
Its $\mathcal{O} \left( (nd)^2 \right)$ complexity is not favourable for scaling in neither instances nor dimensions. 
Exploring avenues for speeding up, possibly along the lines of random features construction, is certainly an important next step in making the algorithm operational for more practical real-life problems.

The method is based on the partial-derivative arguments and therefore assumes the functions (and therefore the kernels) are at least 2nd order differentiable (and square-integrable).
We use here the polynomial and Gaussian kernel as the most commonly used examples.
What other properties of the kernels are necessary to ensure good performance and how the methods could be extended to other, more complex kernels are relevant questions.

The full problem formulation (e.g. equation \eqref{eq:RegLearningEmpiricalH} in proposition \ref{prop:ReprTheorem}) combines the sparse regularizers with the function Hilbert-norm.
This combination has been proposed in \citep{Rosasco2013} to ensure that the regularization part of the problem is strongly convex and the problem is well-posed in terms of the generalization properties.

However, interactions of the Hilbert norm with the sparsity inducing regularizers of section \ref{sec:Regularizers} and the effects on the learning and selection properties are not yet fully clear.
Empirically (from \cite{Rosasco2013} and our own experiments) the models are often little sensitive to variations in $\nu$\footnote{We fix it based on a small subset of replications instead of including it into the full hyper-parameter search.}.

In addition, the $\mcR^{EN}$ regularizer is already strongly convex even without the Hilbert norm.
To what degree combining it with the Hilbert norm is necessary to guarantee good generalization for outside the training needs to be further investigated.
So does its behaviour and the possible improvements it can bring when learning from inputs with non-linear dependencies.
In view of the above considerations, our paper is posing the motivations, foundations and principles for further studies on partial derivative-based regularizations.

\newpage

\bibliographystyle{abbrvnat}
\bibliography{NVSD_UAI2018_v01}

\begin{thebibliography}{24}
\providecommand{\natexlab}[1]{#1}
\providecommand{\url}[1]{\texttt{#1}}
\expandafter\ifx\csname urlstyle\endcsname\relax
  \providecommand{\doi}[1]{doi: #1}\else
  \providecommand{\doi}{doi: \begingroup \urlstyle{rm}\Url}\fi

\bibitem[Argyriou and Dinuzzo(2014)]{Argyriou2014a}
A.~Argyriou and F.~Dinuzzo.
\newblock {A Unifying View of Representer Theorems}.
\newblock In \emph{International Conference on Machine Learning (ICML)}, 2014.

\bibitem[Bach(2009)]{Bach2009a}
F.~Bach.
\newblock {High-Dimensional Non-Linear Variable Selection through Hierarchical
  Kernel Learning}.
\newblock 2009.

\bibitem[Boyd(2010)]{Boyd2010}
S.~Boyd.
\newblock {Distributed Optimization and Statistical Learning via the
  Alternating Direction Method of Multipliers}.
\newblock \emph{Foundations and Trends in Machine Learning}, 2010.

\bibitem[Chan et~al.(2007)Chan, Vasconcelos, and Lanckriet]{Chan2007}
A.~B. Chan, N.~Vasconcelos, and G.~R.~G. Lanckriet.
\newblock {Direct convex relaxations of sparse SVM}.
\newblock \emph{Proceedings of the 24th International Conference on Machine
  Learning (2007)}, 2007.

\bibitem[Chen et~al.(2017)Chen, Stern, Wainwright, and Jordan]{Chen2017}
J.~Chen, M.~Stern, M.~J. Wainwright, and M.~I. Jordan.
\newblock {Kernel Feature Selection via Conditional Covariance Minimization}.
\newblock In \emph{NIPS}, 2017.

\bibitem[Gurram and Kwon(2014)]{Gurram2014}
P.~Gurram and H.~Kwon.
\newblock {Optimal sparse kernel learning in the empirical kernel feature space
  for hyperspectral classification}.
\newblock \emph{IEEE Journal of Selected Topics in Applied Earth Observations
  and Remote Sensing}, 2014.

\bibitem[Hastie et~al.(2015)Hastie, Tibshirani, and Wainwright]{Hastie2015}
T.~Hastie, R.~Tibshirani, and M.~Wainwright.
\newblock \emph{{Statistical Learning with Sparsity: The Lasso and
  Generalizations}}.
\newblock Crc Press, 2015.

\bibitem[Koltchinskii and Yuan(2010)]{Koltchinskii2010}
V.~Koltchinskii and M.~Yuan.
\newblock {Sparsity in multiple kernel learning}.
\newblock \emph{The Annals of Statistics}, 2010.

\bibitem[Lichman(2013)]{Lichman2013}
M.~Lichman.
\newblock {UCI Machine Learning Repository}, 2013.
\newblock URL \url{http://archive.ics.uci.edu/ml}.

\bibitem[Lin and Zhang(2006)]{Lin2006}
Y.~Lin and H.~H. Zhang.
\newblock {Component selection and smoothing in multivariate nonparametric
  regression}.
\newblock \emph{Annals of Statistics}, 2006.

\bibitem[Parikh and Boyd(2013)]{Parikh2013}
N.~Parikh and S.~Boyd.
\newblock {Proximal algorithms}.
\newblock \emph{Foundations and Trends in Optimization}, 2013.

\bibitem[Ravikumar et~al.(2007)Ravikumar, Liu, Lafferty, and
  Wasserman]{Ravikumar2007}
P.~Ravikumar, H.~Liu, J.~Lafferty, and L.~Wasserman.
\newblock {Spam: Sparse additive models}.
\newblock \emph{In Advances in Neural Information Processing Systems}, 2007.

\bibitem[Rosasco et~al.(2013)Rosasco, Villa, and Mosci]{Rosasco2013}
L.~Rosasco, S.~Villa, and S.~Mosci.
\newblock {Nonparametric sparsity and regularization}.
\newblock \emph{The Journal of Machine Learning Research}, 2013.

\bibitem[Saitoh and Sawano(2016)]{Saitoh2016}
S.~Saitoh and Y.~Sawano.
\newblock \emph{{Theory of Reproducing Kernels and Applications}}.
\newblock Springer, 2016.

\bibitem[Sch{\"{o}}lkopf et~al.(2001)Sch{\"{o}}lkopf, Herbrich, and
  Smola]{Scholkopf2001}
B.~Sch{\"{o}}lkopf, R.~Herbrich, and A.~J. Smola.
\newblock {A Generalized Representer Theorem}.
\newblock \emph{COLT/EuroCOLT}, 2001.

\bibitem[Tibshirani(2007)]{Tibshirani1996}
R.~Tibshirani.
\newblock {Regression Shrinkage and Selection via the Lasso}.
\newblock \emph{Journal of the Royal Statistical Society. Series B: Statistical
  Methodology}, 2007.

\bibitem[Tyagi et~al.(2016)Tyagi, Krause, and Eth]{Tyagi2016}
H.~Tyagi, A.~Krause, and Z.~Eth.
\newblock {Efficient Sampling for Learning Sparse Additive Models in High
  Dimensions}.
\newblock \emph{International Conference on Artificial Intelligence and
  Statistics}, 2016.

\bibitem[Weston et~al.(2003)Weston, Elisseeff, Scholkopf, and
  Tipping]{Weston2003}
J.~Weston, A.~Elisseeff, B.~Scholkopf, and M.~Tipping.
\newblock {Use of the Zero-Norm with Linear Models and Kernel Methods}.
\newblock \emph{Journal of Machine Learning Research}, 2003.

\bibitem[Yamada et~al.(2014)Yamada, Jitkrittum, Sigal, Xing, and
  Sugiyama]{Yamada2014}
M.~Yamada, W.~Jitkrittum, L.~Sigal, E.~P. Xing, and M.~Sugiyama.
\newblock {High-dimensional feature selection by feature-wise kernelized
  Lasso.}
\newblock \emph{Neural Computation}, 2014.

\bibitem[Yin et~al.(2012)Yin, Chen, and Xing]{Yin2012}
J.~Yin, X.~Chen, and E.~P. Xing.
\newblock {Group Sparse Additive Models}.
\newblock \emph{International Conference on Machine Learning}, 2012.

\bibitem[Yuan and Lin(2006)]{Yuan2006}
M.~Yuan and Y.~Lin.
\newblock {Model selection and estimation in regression with grouped varibles}.
\newblock \emph{J. R. Statist. Soc. B}, 2006.

\bibitem[Zhao et~al.(2014)Zhao, Li, Liu, and Roeder]{Zhao2014}
T.~Zhao, X.~Li, H.~Liu, and K.~Roeder.
\newblock {CRAN - Package SAM}, 2014.
\newblock URL \url{https://cran.r-project.org/web/packages/SAM/index.html}.

\bibitem[Zhou(2008)]{Zhou2008}
D.~X. Zhou.
\newblock {Derivative reproducing properties for kernel methods in learning
  theory}.
\newblock \emph{Journal of Computational and Applied Mathematics}, 2008.

\bibitem[Zou and Hastie(2005)]{Zou2005}
H.~Zou and T.~Hastie.
\newblock {Regularization and variable selection via the elastic net}.
\newblock \emph{Journal of the Royal Statistical Society. Series B: Statistical
  Methodology}, 2005.

\end{thebibliography}

\newpage

\onecolumn
\appendix

\begin{center}
\toptitlebar
{\Large Structured nonlinear variable selection - supplement}
\bottomtitlebar
\end{center}


\section{Code and replication files}

The implementation of our NVSD algorithm and the replication files for the experiments presented in the main text of our paper are available publicly at the Bitbucket repository \url{https://bitbucket.org/dmmlgeneva/nvsd_uai2018/}. 

\section{Proofs of propositions from the main text}

\begin{proof}[Proof of Proposition 1]
We may decompose any function $f \in \mF$ as $f = f_{\parallel} + f_{\perp}$,
where $f_{\parallel}$ lies in the span of the kernel sections $k_{\bx^i}$ and its partial derivatives $[\partial_a k_{\bx^i}]$ centred at the $n$ training points, and $f_{\perp}$ lies in its orthogonal complement.

The 1st term $\mathcal{\widehat{L}}(f)$ depends on the function $f$ only through its evaluations at the training points $f(\bx^i), i \in \mN_n$.
For each training point $\bx^i$ we have
\begin{equation*}
f(\bx^i) = \langle f, k_{\bx^i} \rangle_\mF = \langle f_{\parallel} + f_{\perp}, k_{\bx^i} \rangle_\mF = \langle f_{\parallel}, k_{\bx^i} \rangle_\mF \enspace ,
\end{equation*}
where the last equality is the result of the orthogonality of the complement $\langle f_{\perp}, k_{\bx^i} \rangle_\mF = 0$.
By this the term $\mathcal{\widehat{L}}(f)$ is independent of $f_{\perp}$.

The 2nd term $\widehat{\mcR}(f)$ depends on the function $f$ only through the evaluations of its partial  derivatives at the training points $\partial_a f(\bx^i), i \in \mN_i, a \in \mN_d$.
For each training point $\bx^i$ and dimension $a$ we have
\begin{equation*}
\partial_a f(\bx^i) = \langle f, [\partial_a k_{\bx^i}] \rangle_\mF = 
\langle f_{\parallel}, [\partial_a k_{\bx^i}] \rangle_\mF 
\enspace ,
\end{equation*}
by the orthogonality of the complement $\langle f_{\perp}, [\partial_a k_{\bx^i}] \rangle_\mF = 0$.
By this the term $\widehat{\mcR}(f)$ is independent of $f_{\perp}$ for the empirical versions of all three considered regularizers $\mcR^L, \mcR^{GL}, \mcR^{EN}$.
For the 3rd term we have $||f||^2_{\mF} = ||f_{\parallel} + f_{\perp}||^2_{\mF} = ||f_{\parallel}||^2_{\mF} + ||f_{\perp}||^2_{\mF}$ because $ \langle f_{\parallel}, f_{\perp} \rangle_{\mF} = 0 $.
Trivially, this is minimised when $f_{\perp} = 0$.
\end{proof}

\begin{proof}[Proof of Proposition 2]
Using the matrices and vector introduced in section 4.1 
and proposition 1
we have
\begin{equation*}\label{eq:fxi}
f(\bx^i) = \sum_{j=1}^n \alpha_j K_{ji} +
\sum_{j=1}^n \sum_{a=1}^d \beta_{aj} \tilde{D}^a_{ij}
\end{equation*}
\begin{equation*}\label{eq:dfxi}
\partial_a f(\bx^i)
=
\sum_{j=1}^n \alpha_j \tilde{D}^a_{ij} +
\sum_{j=1}^n \sum_{c=1}^d \beta_{cj} L^{ca}_{ji}
\end{equation*}

For the 1st term $\mathcal{\widehat{L}}(f)$ we have
\begin{align*}
\mathcal{\widehat{L}}(f) & = \sum_{i=1}^n \big( y^i - f(\bx^i) \big)^2 
 = \sum_{i=1}^n \left( y^i - \sum_{j=1}^n \alpha_j K_{ji} -
\sum_{j=1}^n \sum_{a=1}^d \beta_{aj} \tilde{D}^{a}_{ij} \right)^2 \nn
& =  \sum_{i=1}^n \Big( 
(y^{i})^2 - 2 y^i \sum_{j=1}^n \alpha_j K_{ji} - 2 y^i \sum_{j=1}^n \sum_{a=1}^d \beta_{aj} \tilde{D}^{a}_{ij} 
+ \sum_{j,l}^n \alpha_j \alpha_l K_{ji} K_{l,i}
+ 2 \sum_{j,l}^n \sum_{a=1}^d \beta_{aj} \alpha_l \tilde{D}^{a}_{ij} K_{l,i} \nn
& + \sum_{j,l}^n \sum_{a,b}^d \beta_{aj} \beta_{bl} \tilde{D}^{a}_{ij} \tilde{D}^{b}_{i,l} \Big) \nn
& = 
\by^T \by - 2 \by^T \bK \mathbf{a} - 2 \sum_a^d \by^T \bbD^{a} \bB^T_{a,:} 
+ \pa^T \bK \bK \pa + 2 \sum_a^d \pa^T \bK \bbD^{a} \bB^T_{a,:}
+ \sum_{a,b}^d \bB_{a,:} \bD^a \bbD^{b} \bB^T_{b,:} \nn
& = 
\by^T \by - 2 \by^T \bK \mathbf{a} - 2 \by^T \bD^{T} \pb 
+ \pa^T \bK \bK \pa + 2 \pa^T \bK \bD^{T} \pb 
+ \sum_{a,b}^d \pb ^T \bD \bD^{T} \pb  \nn
& = 
|| \by - \bK \pa - \bD^T \pb ||_2^2  \enspace,
\end{align*}
where $\bB$ is the $d \times n$ matrix with the $\beta$ coefficients $\pb = \vcc({\bB^T})$

For the 2nd term we have
\begin{align*}
\widehat{\mcR}^L(f) & = 
 \sum_{a=1}^d \sqrt{ \frac{1}{n} \sum_{i=1}^n \left( \partial_a f(\bx^i) \right)^2} 
 =  
\sum_{a=1}^d \Bigg[ \frac{1}{n} \sum_{i=1}^n \Big( \sum_{j=1}^n \alpha_j \tilde{D}^{a}_{ji} +
\sum_{j=1}^n \sum_{c=1}^d \beta_{cj} L^{ca}_{ji} \Big)^2 \Bigg]^{0.5} \nn
& =  
\sum_{a=1}^d \Bigg[ \frac{1}{n} \sum_{i=1}^n \Big( 
\sum_{j,l}^n \alpha_j \alpha_l \tilde{D}^{a}_{ji} \tilde{D}^{a}_{l,i} 
+ 2 \sum_{j,l}^n \sum_{c=1}^d \alpha_j \beta_{cl}  \tilde{D}^{a}_{ji} L^{ca}_{l,i} + 
\sum_{j,l}^n \sum_{c,r}^d \beta_{cj} \beta_{rl} L^{ca}_{ji} L^{ra}_{l,i} 
\Big) \Bigg]^{0.5} \nn
& = 
\sum_{a=1}^d \frac{1}{\sqrt{n}} \Bigg[ 
\pa^T \bbD^{a} \bD^{a} \pa +
2 \sum_{c=1}^d \alpha^T \bbD^{a} \bL^{ac} \bB^T_{c:}  
 + \sum_{c,r}^d \bB_{c:} \bL^{ca} \bL^{ar} \bB^T_{r:} 
\Bigg]^{0.5} \nn
& = 
\sum_{a=1}^d \frac{1}{\sqrt{n}} \Bigg[ 
\pa^T \bbD^{a} \bD^{a} \pa +
2 \alpha^T \bbD^{a} \bL^{a} \pb 
 + \pb ^T \bL^{aT} \bL^{a} \pb 
\Bigg]^{0.5} 
 = 
\sum_{a=1}^d \frac{1}{\sqrt{n}} || \bD^a \pa + \bL^a \pb ||_2
\end{align*}

$\widehat{\mcR}^{GL}(f)$ and $\widehat{\mcR}^{EN}(f)$ follow in analogy.

For the 3rd term we have
\begin{align*}
||f||_{\mF}^2 & = || \sum_{j=1}^n \alpha_j k_{\bx^j} + \sum_{j=1}^n \sum_{a=1}^d \beta_{aj} [\partial_a k_{\bx^j}] ||_{\mF}^2 \nn
& = 
\langle \sum_{j=1}^n \alpha_j k_{\bx^j}, \sum_{i=1}^n \alpha_i k_{\bx^i} \rangle_{\mF} 
 + 2 \langle \sum_{j=1}^n \alpha_jk_{\bx^j}, \sum_{i=1}^n \sum_{a=1}^d \beta_{ai} [\partial_a k_{\bx^i}] \rangle_{\mF} \nn
& + \langle \sum_{j=1}^n \sum_{a=1}^d \beta_{aj} [\partial_a k_{\bx^j}], \sum_{i=1}^n \sum_{c=1}^d \beta_{ci} [\partial_c k_{\bx^i}] \rangle_{\mF} \nn
& = 
\pa^T \bK \pa + 
2 \sum_{ij}^n \sum_a^d \alpha_j \beta_{ai} \, \partial_a k_{\bx^j}(\bx^i) 
 + \sum_{ij}^n \sum_{ac}^d \beta_{aj} \beta_{ci} \frac{\partial^2}{\partial x^j_a \partial x^i_c} k(\bx^j,\bx^i) \nn
& = 
\pa^T \bK \pa + 
2 \sum_{ij}^n \sum_a^d \alpha_j \beta_{ai} \tilde{D}_{ji}^{a}
+ \sum_{ij}^n \sum_{ac}^d \beta_{aj} \beta_{ci} L_{ji}^{ac}  \nn
& = 
\pa^T \bK \pa + 
2 \sum_a^d \pa^T \bbD^{a} \bB^T_{a:} 
+ \sum_{ac}^d \bB_{:j} \bL^{ac} \bB^T_{c:} \nn
& = 
\pa^T \bK \pa + 
2 \pa^T \bD^{T} \pb 
+ \sum_{a}^d \bB_{a:} \bL^{a} \pb \nn
& = 
\pa^T \bK \pa + 
2 \pa^T \bD^{T} \pb 
+ \pb ^T \bL \pb 
\end{align*}
\end{proof}

\begin{proof}[Proof of Proposition 4]
The proximal problem in step $S2$ for $\mcR^L$ for a single partition $\pv_a$ is
\begin{equation*}
\mcR^L: \ \pv_a^{(k+1)} =  
\argmin_{\pv_a} \frac{\tau}{\sqrt{n}}  ||\pv_a||_2 + \frac{\rho}{2} ||\bZ^a \, \po^{(k+1)} - \pv_a + \pl_a^{(k)} ||_2^2
\end{equation*}

This convex problem is non-differentiable at the point $\pv = \vc{0}$.
It is, however, sub-differentiable with the optimality condition for the minimizing $\pv^*$
\begin{equation*}
\vc{0} \in \partial \, \frac{\tau}{\sqrt{n}}  ||\pv^*_a||_2 - \rho \, (\bZ^a \, \po^{(k+1)} - \pv_a + \pl_a^{(k)}) \enspace ,
\end{equation*}
where for any function $f : \mR^d \to \mR$, $\partial f(\bx) \subset \mR^d$ is the sub-differential of $f$ at $x$ defined as
\begin{equation*}
\partial f(\bx) = \{ \bg \, | \, f(\bz) \ge f(\bx) + \bg^T (\bz - \bx) \} \enspace .
\end{equation*}

For notational simplicity, in what follows we introduce the variable $\bv = \bZ^a \, \po^{(k+1)} + \pl_a^{(k)}$, and we drop the sub-/super-scripts of the partitions $a$ and the iterations $k$.

\paragraph{Part A} For all points other than $\pv^* = \vc{0}$ the optimality condition reduces to 
\begin{equation*}
\vc{0} = \frac{\tau}{\sqrt{n}} \, \frac{\pv^*}{||\pv^*||_2} - \rho \, (\bv - \pv^*) \enspace ,
\end{equation*}
From which we get
\begin{eqnarray*}
\left( \frac{\tau}{\rho \sqrt{n}||\pv^*||_2} + 1  \right) \pv^* & = & \bv \\
\left( \frac{\tau}{\rho \sqrt{n}||\pv^*||_2} + 1  \right) || \pv^* ||_2 & = & ||\bv||_2 \\
|| \pv^* ||_2 & = & ||\bv||_2 - \frac{\tau}{\rho \sqrt{n}} \enspace .
\end{eqnarray*}
We use this result in the optimality condition 
\begin{eqnarray*}
\vc{0} & = & \frac{\tau}{\sqrt{n}}  \frac{\pv^*}{||\bv||_2 - \frac{\tau}{\rho \sqrt{n}}} - \rho \, (\bv - \pv^*) \\
\frac{\tau}{\sqrt{n}} \, \pv^* & = & \rho \, (\bv - \pv^*) (||\bv||_2 - \frac{\tau}{\rho \sqrt{n}}) \\
\frac{\tau}{\sqrt{n}} \, \pv^* & = & (\rho ||\bv||_2 - \frac{\tau}{\sqrt{n}}) \bv 
- \rho \, ||\bv||_2 \, \pv^* + \frac{\tau}{\sqrt{n}} \, \pv^* \\
\pv^* & = & \left( 1 - \frac{\tau}{ \rho \sqrt{n}||\bv||_2} \right) \bv
\end{eqnarray*}

\paragraph{Part B} For the point $\pv^* = \vc{0}$ we have $\partial ||\pv^*||_2 = \{ \bg \, | \, ||\bg||_2 \leq 1 \}$ (from the definition of sub-differential and the Cauchy-Schwarz inequality).

From the optimality condition 
\begin{eqnarray*}
\vc{0} & = & \frac{\tau}{\sqrt{n}} \, \bg - \rho \, \bv  \qquad \qquad (\pv^* = \vc{0}) \\
\rho \, \bv & = & \frac{\tau}{\sqrt{n}} \, \bg \\
\rho \, ||\bv||_2 & = & \frac{\tau}{\sqrt{n}} \, ||\bg||_2 \\
||\bv||_2 & \leq & \frac{\tau}{\rho \sqrt{n}} \qquad \qquad (||\bg||_2 \leq 1) \\
\end{eqnarray*}

Putting the results from part A and B together we obtain the final result
\begin{equation*}
\pv^* = \left( 1 - \frac{\tau}{ \rho \sqrt{n}||\bv||_2} \right)_+ \bv
\end{equation*}

The proofs for $\mcR^{GL}$ and $\mcR^{EN}$ follow similarly.
\end{proof}

\section{Examples of kernel partial derivatives}

We list here the 1st and 2nd order partial derivatives which form the elements of the derivative matrices $\bD$ and $\bL$ introduced in section 4.1 for some common kernel functions $k$.

\paragraph{Linear kernel} \ \\ 
Kernel gram matrix
\begin{align*}
K_{i,j} = k(\bx^i,\bx^j) = \langle \bx^i, \bx^j \rangle
\end{align*}

1st order partial-derivative matrix
\begin{align*}
D^a_{i,j} = \frac{\partial k(\bs,\bx^j)}{\partial s_a}|_{\bs=\bx^i} =  x^j_a
\end{align*}

2nd order partial-derivative matrix
\begin{align*}
L^{ab}_{i,j} = \frac{\partial^2 k(\bs,\mathbf{r})}{\partial s_a \partial r_b}|_{{\bs=\bx^i}\atop{\mathbf{r}=\bx^j}}
 = 
\begin{cases}
0 & \text{if } a \neq b \\
1 & \text{if } a = b \\
\end{cases}
\end{align*}

\paragraph{Polynomial of order $p>1$} \ \\
Kernel gram matrix
\begin{align*}
K_{i,j} = (\langle \bx^i, \bx^j \rangle + c)^p
\end{align*}

1st order partial-derivative matrix
\begin{align*}
D^a_{i,j} = p \, (\langle \bx^i, \bx^j \rangle + c)^{p-1} \ x^j_a
\end{align*}

2nd order partial-derivative matrix
\begin{align*}
L^{ab}_{i,j} = 
\begin{cases}
p (p-1) \, (\langle \bx^i, \bx^j \rangle + c)^{p-2} \ x^i_b x^j_a \qquad \text{if } a \neq b \\
p (p-1) \, (\langle \bx^i, \bx^j \rangle + c)^{p-2} \ x^i_a x^j_a + p \, (\langle \bx^i, \bx^j \rangle + c)^{p-1} \\
\qquad \qquad \qquad \qquad  \qquad \qquad \qquad \qquad \text{if } a = b \\
\end{cases}
\end{align*}

\paragraph{Gaussian kernel} \ \\
Kernel gram matrix
\begin{align*}
K_{i,j} = \exp \left( - \frac{||\bx^i - \bx^j||_2^2}{2\sigma^2} \right) 
\end{align*}

1st order partial-derivative matrix
\begin{align*}
D^a_{i,j}  = \exp \left( - \frac{||\bx^i - \bx^j||_2^2}{2\sigma^2} \right) \frac{x^j_a - x^i_a}{\sigma^2}
\end{align*}

2nd order partial-derivative matrix
\begin{align*}
L^{ab}_{i,j} = 
\begin{cases}
\exp \left( - \frac{||\bx^i - \bx^j||_2^2}{2\sigma^2} \right) \frac{(x^j_a - x^i_a)(x^i_b - x^j_b)}{\sigma^4}
 & \text{if } a \neq b \\
\exp \left( - \frac{||\bx^i - \bx^j||_2^2}{2\sigma^2} \right) \frac{(x^i_a - x^j_a)^2 - \sigma^2}{-\sigma^4}
 & \text{if } a = b \\
\end{cases}
\end{align*}


\end{document}